\documentclass[letterpaper, 10 pt, conference]{ieeeconf}

\newif\ifarxiv
\arxivtrue    % Activate arxiv mode
\makeatletter
\let\NAT@parse\undefined
\makeatother

\usepackage{bm}
\usepackage{url}
\usepackage{verbatim}
\usepackage{graphicx}
\usepackage{cite}
\usepackage{lipsum}
\usepackage{amsmath}
\usepackage{amssymb}
\usepackage{nccmath}
\usepackage{comment}
\usepackage{balance}
\usepackage{multirow}
\usepackage{textcomp}
\usepackage{multirow}
\usepackage[caption=false,font=footnotesize]{subfig}
\usepackage{algorithm}
\usepackage{dblfloatfix}
\usepackage[table]{xcolor}
\usepackage[noend]{algpseudocode}
\usepackage{booktabs} % methods for table
\usepackage{xspace} % for xspace
\usepackage{gensymb} % methematical variables (e.g. \degree)
\usepackage{times} % \emph{} to underline-like behavior, No.1
\usepackage{bm} % \emph{} to underline-like behavior, No.2
\usepackage{mathtools} % for mathematical unit and tools (e.g. \lbrack)
\usepackage{siunitx}
\usepackage{lineno} % For lineno, but FAIL when conference template
\usepackage{tcolorbox}

\definecolor{cvprblue}{rgb}{0.21,0.49,0.74}
\usepackage[pagebackref=true,breaklinks=true,letterpaper=true,colorlinks,bookmarks=false,citecolor=cvprblue]{hyperref} 

\newcommand{\etal}{\emph{et al.}\xspace}

\newcommand{\Acronym}[0]{SG-Init\xspace} 
\newcommand{\AcroAcro}[0]{SG-Init+D\xspace} 
\newcommand{\AcroAcroNos}[0]{SG-Init+D} 
\newcommand{\AcronymSLAM}[0]{SG-Init + DROID\xspace} 
\newcommand{\zeroshot}{zero-shot\xspace}
\newcommand{\sota}{state-of-the-art\xspace}

\def\blue#1{\textcolor{blue}{#1}}
\definecolor{darkgreen}{rgb}{0.0, 0.5, 0.0}
\newcommand{\darkgreen}[1]{\textcolor{darkgreen}{#1}}
\def\orange#1{\textcolor{orange}{#1}}

\newlength\dunder
\begin{document}

\title{Self-Supervised Geometry-Guided Initialization \\ for Robust Monocular Visual Odometry}

\author{Takayuki Kanai$^{1}$,
  Igor Vasiljevic$^{2}$,
  Vitor Guizilini$^{2}$,
  and Kazuhiro Shintani$^{1}$%
  \thanks{$^{1}$T. Kanai and K. Shintani are with Frontier Research Center, Toyota Motor Corporation (TMC) in Toyota, Aichi, Japan. {\tt\footnotesize \{first\_lastname\}@mail.toyota.co.jp}}% <-this % stops a space
  \thanks{$^{2}$I. Vasiljevic and V. Guizilini are with Toyota Research Institute (TRI) in Los Altos, California, United States. {\tt\footnotesize \{first.lastname\}@tri.global}}% <-this % stops a space
}

\makeatletter
\g@addto@macro\@maketitle{
  \begin{figure}[H]
    \setlength{\linewidth}{\textwidth}
    \setlength{\hsize}{\textwidth}
    \centering
    \includegraphics[width=.94\linewidth]{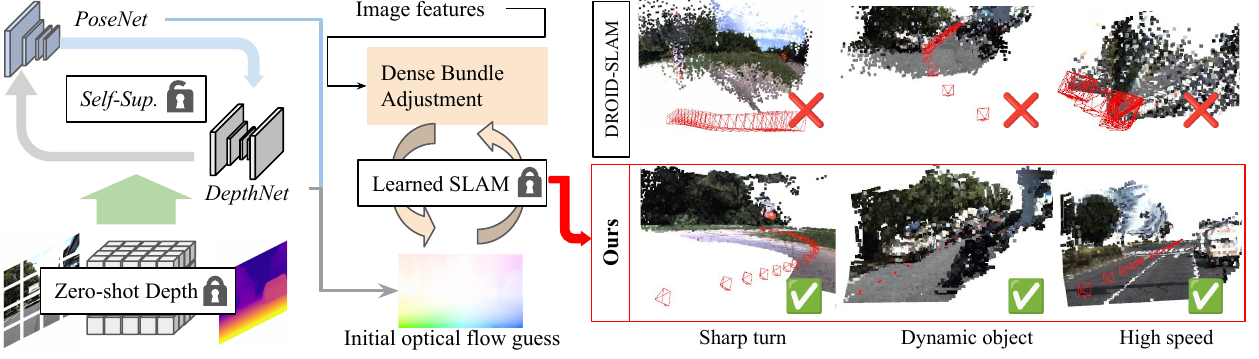}
    % \vspace{-5mm}
    \caption{Diagram showing the proposed method (left) alongside our \Acronym results (bottom right) compared with the baseline (top right). Although large amounts of camera motion, including dynamic environments, pose a challenge to previous learning-based SLAM~\cite{teed2021droid}, we achieve significant improvements owing to the proposed novel initialization scheme enabled by self-supervised geometric priors (\textit{DepthNet} and \textit{PoseNet}), with a zero-shot depth training.}
    \label{fig:teaser}
    \vspace{-8mm}
  \end{figure}
}
\makeatother

\setcounter{figure}{0}
\renewcommand{\thesubfigure}{\Alph{subfigure}}

\maketitle

\begin{abstract}
  Monocular visual odometry is a key technology in various autonomous systems.
Traditional feature-based methods suffer from failures due to poor lighting, insufficient texture, and large motions.
In contrast, recent learning-based dense SLAM methods exploit iterative dense bundle adjustment to address such failure cases, and achieve robust and accurate localization in a wide variety of real environments, without depending on domain-specific supervision.
However, despite its potential, the methods still struggle with scenarios involving large motion and object dynamics.
In this study,
we diagnose key weaknesses in a popular learning-based dense SLAM model (DROID-SLAM) by analyzing major failure cases on outdoor benchmarks and exposing various shortcomings of its optimization process.
We then propose the use of self-supervised priors leveraging a frozen large-scale pre-trained monocular depth estimator to initialize the dense bundle adjustment process, leading to robust visual odometry without the need to fine-tune the SLAM backbone.
Despite its simplicity, the proposed method demonstrates significant improvements on KITTI odometry, as well as the challenging DDAD benchmark.
The project page:~\url{https://toyotafrc.github.io/SGInit-Proj/}
% (174/200 words)
\end{abstract}

\addtocounter{footnote}{+1}
\footnotetext{T. Kanai and K. Shintani are with Frontier Research Center, Toyota Motor Corporation (TMC), in Toyota, Aichi, Japan.
    {\tt\footnotesize \{first\_lastname\}@mail.toyota.co.jp}}
\addtocounter{footnote}{+1}
\footnotetext{I. Vasiljevic and V. Guizilini are with Toyota Research Institute (TRI), in Los Altos, California, United States. {\tt\footnotesize \{first.lastname\}@tri.global}}
\addtocounter{footnote}{-2}
\vspace{.5em}
\IEEEoverridecommandlockouts

\IEEEpeerreviewmaketitle
\setcounter{figure}{1}

%%%%%%%%%%%%%%%%%%%%%%%%%%%%%%%%%%%%%%%%%%%%%%%%%%%%%%%
\section{Introduction}
Visual odometry, a special case of simultaneous localization and mapping (SLAM), is a fundamental task in the development of autonomous systems.
This is demonstrated by the proliferation of several potential applications~\cite{xie2022neural,zhou2023nerf,neo2023iccv}.
For this new challenge, data-driven deep-learning-based methods have garnered significant attention because of their robustness and higher accuracy relative to traditional methods~\cite{tang2020kp3d,dfvo2020icra,pan2020pseudoegbd}. In particular, a series of strategies that exploit dense bundle adjustment~\cite{teed2021droid,Ye2023PVO,hagemann2023droidcarib,li2024_MegaSaM} exhibit significant improvement without requiring domain-specific supervision, by leveraging the learned knowledge from synthetic datasets~\cite{tartanair2020iros}.

\begin{figure}[t!]
	\vspace{2mm}
	\begin{center}
		\includegraphics[width=0.47\textwidth]{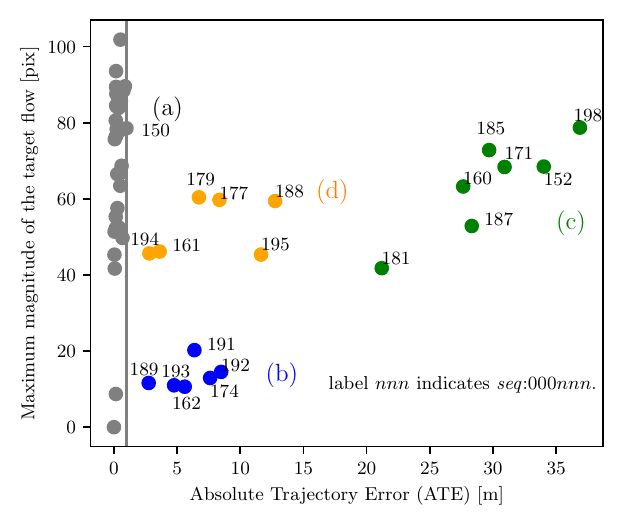}
	\end{center}
	\vspace{-5mm}
	\caption{\textbf{Failure mode analysis of DROID-SLAM~\cite{teed2021droid} on DDAD \textit{val.} sequences~\cite{packnet}.} 
        Categories are defined by (1) thresholding ATE at $1.0$~\lbrack \si{\metre}\rbrack \ for success/failure and (2) manual clustering. 
        We observed that the failure mode splits into: (b) insufficient motion, (c) highly dynamic ego-motion, and (d) where patterns of (b) and (c) are included.
		Please refer to Section~\ref{subsec:ddad} for per-category analysis.
	}
	\label{fig:ddad_corr}
	\vspace{-2em}
\end{figure}

Nevertheless, the \textit{off-the-shelf} use of its pre-trained model is prone to deterioration in driving scenarios\cite{Ye2023PVO,r3d3,yin2023metric} when compared with traditional methods (Fig.~\ref{fig:teaser}).
Notably, even with domain-specific fine-tuning of the \textit{backbone}---a common strategy in domain adaptation---performance does not always improve on the target domain~\cite{r3d3}. 
In addition, any fine-tuning of the backbone will specialize the SLAM system for a particular setting, resulting in the loss of generality. Thus, rather than modifying the backbone, we seek to understand how optimization failure causes deterioration in these settings.
In fact, our analysis suggests that these methods struggle with large ego movements the most severely in their earlier timesteps\footnote{Defined by a forward motion of $15.0$\si{\metre}, where keyframes for bundle adjustment are insufficiently recorded in \cite{teed2021droid}.} (Fig.~\ref{fig:ddad_corr}, category (c)).
These scenarios require a large displacement of optical flow estimation from a poor initialization
(e.g., the situation in which the `initial optical flow guess' is not obtained in Figure~\ref{fig:teaser}).
Therefore, optimization is prone to converging quickly into an inaccurate solution.
Additionally, we also observe the presence of dynamic objects and a large ratio of textureless areas causes sub-optimal initialization.

In this study, we investigate learning-based SLAM as an optimization problem, and propose a novel strategy designed to increase robustness and stability by addressing failure cases due to poor optimization convergence.
First, we experimentally demonstrate various vulnerabilities of the dense optimization process inside the learned SLAM module. Then, we show the alleviation of this issue by geometric prior guidance during initialization, according to the principle of self-supervised depth and ego-motion learning.
Furthermore, we show the benefits of using large-scale pre-trained \textit{\zeroshot} depth estimators to guide the self-supervision process toward fully leveraging the \textit{Dense} learned backbone.
This \zeroshot guidance alleviates the inherent ambiguity of self-supervised training, which is crucial for not feeding the degraded depth into the dense backbone ---an \textbf{all}-pixel-processing optimizer.
Consequently, the proposed method efficiently enhances SLAM performance without requiring any modification to its pre-trained weights. We demonstrate this by achieving significant improvements on the challenging DDAD~\cite{packnet} benchmark, comprising diverse driving scenes, as well as on the standard KITTI~\cite{geiger2012we} benchmark.

In summary, we propose \textbf{\Acronym} (\textbf{S}elf-Supervised \textbf{G}eometry-Guided \textbf{Init}ialization), a novel method designed to robustify visual odometry through geometric initialization.
Our contributions are as follows:

\begin{itemize}
    \item We expose major weaknesses of learning-based SLAM strategies that rely on dense bundle adjustment, and propose a novel method, \textbf{Self-Supervised Geometric-Guided Initialization (\Acronym)}, to improve visual odometry performance under challenging conditions.
    \item We experimentally demonstrate that off-the-shelf \zeroshot monocular depth estimators ~\cite{Wei2021CVPRleres,kar20223d} can be integrated into the proposed self-supervised framework to \textbf{further boost \textit{dense} SLAM performance}.
    \item We provide a \textbf{comprehensive analysis of our proposal adapted to DROID-SLAM~\cite{teed2021droid}} on the standard KITTI benchmark~\cite{geiger2012we}, as well as challenging driving scenes from DDAD~\cite{packnet}, and provide insights into how to construct future visual SLAM systems.

\end{itemize}

%%%%%%%%%%%%%%%%%%%%%%%%%%%%%%%%%%%%%%%%%%%%%%%%%%%%%%%
\section{Related Work}
\noindent\textbf{Visual Odometry.}
A previous common approach to visual odometry is \textit{indirect-and-sparse}~\cite{Engelpami2018DSO}:
handcrafted feature extraction and matching on images were essential to its initial success~\cite{murORB2,orb3}.
% Contrary, 
Studies revealed that introducing learning-based priors led to more geometrically consistent estimations~\cite{bescos2018dynaslam,bian2021ijcv,pan2020pseudoegbd}.
However, the use of sparse descriptors limited the amount of information compared with the dense inputs.
Thus, these approaches risked large estimation errors or complete failures~\cite{tri_sesc_iros23}.

%%% shortened to follow the reviewer's comment
On the contrary, strategies that preserve the density of the raw input to gain accuracy and robustness have also been studied~\cite{teed2021droid,d3vo2020cvpr,tang2020kp3d,dfvo2020icra,pan2020pseudoegbd,Ye2023PVO,hagemann2023droidcarib,li2024_MegaSaM,r3d3}.
One of their core strategies is dense bundle adjustment with large-scale pre-training, which originated from DROID-SLAM~\cite{teed2021droid}.
The present study is also categorized into this branch, which addresses the typical deterioration of the dense bundle adjustment without additional supervised training.
Notably, as our proposal addresses a fundamental step for the dense bundle adjustment, it can be directly integrated into several different frameworks~\cite{Ye2023PVO,r3d3,hagemann2023droidcarib,li2024_MegaSaM} beyond DROID-SLAM.

\def\red#1{\textcolor{red}{#1}}

%%% Add discussion about hardware, and the other SLAM initilization 
\noindent\textbf{SLAM Initialization with Self-supervised Priors.}
This study focuses on self-supervised depth and ego-motion learning to initialize the input for the SLAM module ~\cite{zhou2017unsupervised,godard2017unsupervised}.
The self-supervision provides interframe-consistent depth and pose predictions by videos, which can be seamlessly adapted to the SLAM systems~\cite{tri_sesc_iros23,pan2020pseudoegbd,bian2021ijcv,dfvo2020icra,tang2020kp3d,d3vo2020cvpr}, across various camera models
~\cite{tri-selfcalibration}.
Hence, once the priors are trained, our initialization strategy is expected to be domain-agnostic, time-efficient, and can easily be integrated with typical initializations, such as M-estimation~\cite{hu2013reliable} or a spectral-based initialization~\cite{doherty2022initialization}.

%%% Redesigning the structure (keep the story tho
Despite those attractive features, naive integration of the self-supervised networks into the dense bundle adjustment poses one key difficulty: the local degradation of the depth estimation.
Although this degradation is inevitable owing to inherent shortcomings of self-supervised learning, the dense bundle adjustment requires \textbf{all} pixels as input.
Thus, globally reliable dense depth prediction is required for the dense bundle adjustment, unlike Pseudo-RGBD SLAM~\cite{pan2020pseudoegbd} or KP3D~\cite{tang2020kp3d}, where the sparse SLAM backbone leverages the local pixels of depth prediction.
Conversely, leveraging all pixels and adaptively adjusting the weight of nonlinear optimization across broader temporal context (unlike \textit{uncertainty}~\cite{d3vo2020cvpr} or \textit{score}~\cite{tang2020kp3d}) are expected to result in improvements over sparse-based strategies~\cite{pan2020pseudoegbd,bian2021ijcv}.

Consequently, we also propose leveraging the capabilities of a zero-shot depth estimation model to provide a more accurate dense depth estimation as an initializer for the dense bundle adjustment.

\noindent\textbf{Zero-shot Monocular Depth Estimation.}
Given the ever-increasing availability of internet-wide data, computation power, and more efficient training algorithms, a wide range of monocular depth estimators have been proposed~\cite{MDLi18,Wei2021CVPRleres,kar20223d,yin2023metric,tri-zerodepth}.
These models hold a remarkable \zeroshot capability by fully leveraging the labeled supervising dataset, including in the self-supervised training objective.
Although the methods are more robust to self-supervised models, various challenges face in their downstream adoption: (1) efficiency, both in terms of memory footprint and inference speed; and (2) difficulty in achieving interframe scale consistency~\cite{depthanything}, although several studies have addressed to this problem~\cite{yin2023metric,tri-zerodepth}.
We operationally address these challenges by leveraging two sources of geometric priors simultaneously: zero-shot \emph{relative} depth estimators (scale awareness is not mandatory), and the self-supervised estimators (interframe-consistency is promised) of a relatively lighter model.

%%%%%%%%%%%%%%%%%%%%%%%%%%%%%%%%%%%%%%%%%%%%%%%%%%%%%%%
\section{Methodology}
\begin{figure*}[t]
    \vspace{2mm}
    \begin{center}
        \includegraphics[width=0.90\textwidth]{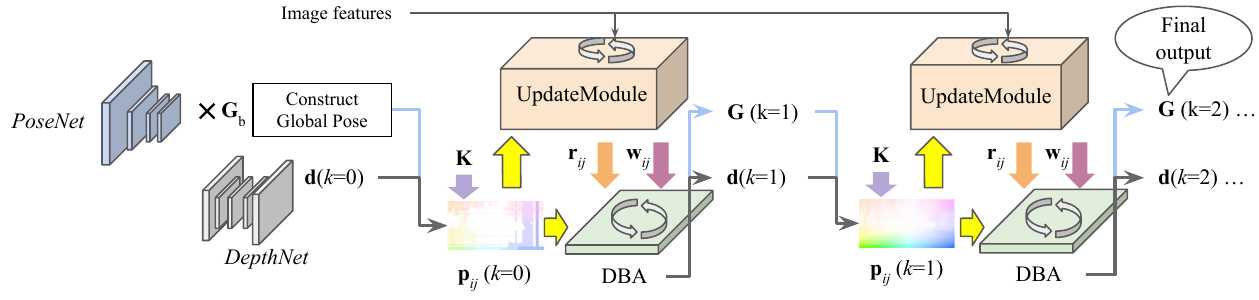}
    \end{center}
    \vspace{-3mm}
    \caption{\textbf{\Acronym for visual odometry stage}. The learned geometric priors, \textit{DepthNet} and \textit{PoseNet} from self-supervised learning provide the initial guess for both \textit{UpdateModule} and dense bundle adjustment (\textit{DBA}) layer, combined with the camera intrinsics $\mathbf{K}$. Then the target variables are recursively updated to obtain the final output.}
    \label{fig:sd3_with_dba}
    \vspace{-5mm}
\end{figure*}
\subsection{Preliminaries}
\label{sec:mono}
\noindent\textbf{Dense Bundle Adjustment and UpdateModule.}
In the dense bundle adjustment step, a series of inverse depth ($\mathbf{d}$) and camera pose ($\mathbf{G}$) are refined recursively guided by the confidence estimation.
Given a pair of images $(I_{i}, I _{j})$ that share co-visible area, the pre-trained operator (modeled by a neural network parameterized with $\theta$) predicts the tuple of the revision of the optical flow $\mathbf{r}_{ij}(k)$ and its associated weight, based on the currently estimated flow $\mathbf{p}_{ij}(k)$:
\begin{equation}
\small
  \mathbf{r}_{ij}(k)
  , \mathbf{w}_{ij}(k)
  =
  \textit{UpdateModule}_{\theta}(I_{i}, I_{j}, \mathbf{p}_{ij}(k))
  \label{eq:update_module}
\end{equation}

\noindent where $k$ indicates the iteration step of the recurrent neural network model, which we will omit for simplicity unless needed.
Note that $\mathbf{p}_{ij}$ is obtained from the following operation given the camera intrinsics $\mathbf{K}$ (and, $\mathbf{K}$ is also subsequently omitted unless needed):
\begin{equation}
  % \small
  \mathbf{p}_{ij}
  =
  \Pi_{c} (\mathbf{G}_{ij} \circ \Pi_{c}^{-1}(\mathbf{p}_i, \mathbf{d}_i | \mathbf{K}) | \mathbf{K})
  \label{eq:fwd_flow}
\end{equation}

\noindent where $\textbf{p}_{i}$ denotes the grid pixel on the image coordinate of $I_{i}$, $\textbf{d}_{i}$ denotes its inverse depth prediction, $\mathbf{G}_{ij}$ denotes the relative camera transformation calculated by $\mathbf{G}_{j} \circ \mathbf{G}^{-1}_{i}$, $\Pi_{c}$ denotes projection operator given the camera model, and $\Pi_{c}^{-1}$ denotes that of ``inverse'' operation (lifting-up to 3D point cloud).

Next, the residual between the “tentative” target of the flow $\mathbf{p}_{ij}^*=\mathbf{r}_{ij}+\mathbf{p}_{ij}$ and the warped pixel of frame $i$ to $j$ (Eqn.~\ref{eq:fwd_flow}) is minimized in the dense bundle adjustment layer:

\begin{equation}
  % \small
  E (\mathbf{G}, \mathbf{d}) =
  \sum_{(i,j) \in \mathcal{E}} \left\| \mathbf{p}_{ij}^* - \Pi_{c} (\mathbf{G}_{ij} \circ \Pi_{c}^{-1}(\mathbf{p}_i, \mathbf{d}_i)) \right\|^2_{\Sigma_{ij}}
  \\
  \quad
  \label{eq:dense_ba}
\end{equation}

\noindent where $\Sigma_{ij} = \text{diag} \lbrack \mathbf{w}_{ij} \rbrack$, the indices $(i,j) \in \mathcal{E}$ describe the co-visible keyframe pairs, and
$\left\| \cdot \right\|_{\Sigma_{ij}}$ is the Mahalanobis distance defined by the predicted weights $\mathbf{w}_{ij}$.
Because the minimization is formulated as a linear problem similar to a standard SLAM formulation, it can be efficiently solved by the extension of the Gauss--Newton Step~\cite{teed2021droid}.

Notably, this optimization requires an initial depth and pose guess for the nonlinear optimization.
Generally, an incorrect or noisy initialization will lead to a local minimum~\cite{hu2013reliable,doherty2022initialization}. 
Moreover, we experimentally demonstrate that it indeed causes erroneous localization, particularly when a large displacement of the flow is required, or a moving object is observed.

%%% A bit rephrased
\noindent\textbf{Self-Supervised Depth and Ego-motion Modeling.}
Self-supervised depth and ego-motion learning is formulated as the joint optimization of depth and pose neural networks (henceforth referred to as \textit{DepthNet} and \textit{PoseNet}).
The photometric loss $\mathcal{L}_{p}$ propagates the learning signal through the entire architecture:
\begin{equation}
  \small
  \mathcal{L}_{p}(I_t,\hat{I_t}) = \alpha~\frac{1 - \text{SSIM}(I_t,\hat{I_t})}{2} + (1-\alpha)~\| I_t - \hat{I_t} \|
  \label{eq:photo_mono}
\end{equation}

\noindent where $I_{t}$ denotes a target image, and $\hat{I_{t}}$ is an image synthesized via the photometric warping operation~\cite{jaderberg2015spatial}, obtained by the four variables: (1) context image $I_{c}$, (2) predicted depth $\hat{D}_{t}$ from \textit{DepthNet}, (3) predicted extrinsics $\hat{\mathbf{X}}^{t \to c}~\in~\text{SE(3)}$ induced by the ego-motion from \textit{PoseNet}, and (4) camera intrinsics.
Note that the ill-posedness of photometric-based optimization, stemming from dynamic observation, and luminance shift, causes multiple sources of inaccuracies in this learning.
We thus tackle this by leveraging the \zeroshot capabilities of a large-scale pre-trained model while ensuring its scale consistency.

\subsection{Dense Bundle Adjustment with Geometric Initialization}
\label{sec:pcc++}

\noindent \textbf{Self-Supervised Geometry-Guided Initialization}. 
We introduce a novel approach that leverages the self-supervised geometric priors to properly initialize the SLAM system~(Fig.~\ref{fig:sd3_with_dba}).
Accordingly, we first obtain the geometric priors: \textit{DepthNet} and \textit{PoseNet} by self-supervised learning (details are described later).
Next, we use these prior predictions to compose the initial flow $\mathbf{p}_{ij}(k=0)$, and bridge to the \textit{UpdateModule} (Eqn.~\ref{eq:update_module}) and the dense bundle adjustment layer (Eqn.~\ref{eq:dense_ba}).
In particular, (1) $\textbf{d}_{i}$ is naively obtained from the inverse of \textit{DepthNet} prediction as $\hat{D}^{-1}_{i}$,
and (2) the camera pose $\mathbf{G}_{i}$ is calculated by chaining the relative pose estimation from the \textit{PoseNet} predictions $\hat{\mathbf{X}}^{\tau \to \tau+1}$ as:

\begin{equation}
  \small
  \mathbf{G}_{i} = \hat{\mathbf{X}}^{i-1 \to i} \cdot \cdot \cdot \xspace \hat{\mathbf{X}}^{\text{b} + 1 \to \text{b} + 2} \circ \hat{\mathbf{X}}^{\text{b} \to \text{b} + 1} \circ \mathbf{G}_{\text{b}}
  \label{eq:camera_chain}
\end{equation}

\noindent 
where $\mathbf{G}_{\text{b}}$ denotes the camera pose of the base frame to start the transformation chaining.
Providing more accurate initial guesses is expected to let the \emph{UpdateModule} behave similar to its training phase, in terms of (1) a first few keyframe poses are relatively accurate~\cite{teed2021droid}, and (2) the optimization is started from a coarsely estimated flow~\cite{teed2020raft}. 
Therefore, our proposal can be rephrased as reducing the train/test \emph{domain-gap} of the behavior of the learned module, rather than offering in-domain supervision to reduce it.
Although the \textit{UpdateModule} still relies on input images $I_{i}$ and $I_{j}$ --- implying visual domain shifts such as synthetic-to-real cannot be entirely addressed ---, we demonstrate that aligning the geometric aspects during training and testing suffices to achieve significant performance improvements.

In the following experiment, we set $\text{b}=0$ (thus, $\mathbf{G}_{\text{b}}$ is always the origin of the odometry), and use \textit{PoseNet} only for the bundle adjustment to the first constructed keyframe pairs (i.e., $\mathbf{G}_{0}, \mathbf{G}_{1}, ... \mathbf{G}_{7}$).
Subsequently, the camera poses~$\mathbf{G}_{ij}(k=0)$ is provided as a way of the original DROID-SLAM~\cite{teed2021droid}.

\noindent \textbf{Zero-shot Depth-Guidance for Prior Learning}.
We address the inherent ambiguity of self-supervised depth and ego-motion learning by also leveraging zero-shot monocular depth estimation. 
In particular, we employ zero-shot pseudo-labeling as another source of supervision to tackle the inherent problems of self-supervision, such as infinite depth predictions for dynamic objects~\cite{packnet-semguided} and edge bleeding~\cite{monodepth2}.
We applied the vanilla-learning scheme of SC-Depth V3 to validate the improvements produced by this approach~\cite{sc_depthv3_tpami_2024}, as it provides a naive detector for ill-posedness, which is expected to violate the dense depth prediction and therefore the downstream task of optical flow estimation.
However, other knowledge-distillation strategies can also be applied~\cite{wang2024altnerf}.

%%%%%%%%%%%%%%%%%%%%%%%%%%%%%%%%%%%%%%%%%%%%%%%%%%%%%%%
\section{Experiments}
\label{sec:experiment}

%%% Explain more about the metric
We used common metrics to comprehensively study our proposal and compare it against the relevant baselines.
For odometry, we employed the absolute trajectory error with ground truth (GT) alignment, defined as:
\begin{equation}
    \small
    \text{ATE}: \sqrt{\frac{1}{T} \sum_{i=0}^{T-1} \| \text{trans}(\mathbf{Q}_i^{-1} \mathbf{S} \mathbf{P}_i) \|^2}
    \label{eq:ATE}
\end{equation}
where $\mathbf{S}$ denotes the alignment transformation applied to the predicted trajectory $\mathbf{P}_{i}$ to best fit the GT trajectory $\mathbf{Q}_i$, and $T$ denotes the number of camera frames.
Depending on the benchmark, we show two types of this metric: (1) ATE, where a uniform scaling transformation is assigned to $\mathbf{S}$, and (2) ATE$\dagger$, which employs a full similarity transformation to obtain $\mathbf{S}$.
These transformations are determined to minimize the alignment error between the predicted and the GT trajectory~\cite{umeyama1991}.

%%%%%%%%%%%%%%% if Abs.Rel. is also needed as equation, use this %%%%%%%%%
% \begin{equation}
%     \footnotesize
%     \text{ATE}: \sqrt{\frac{1}{T} \sum_{i=1}^{T} \| \text{trans}(\mathbf{Q}_i^{-1} \mathbf{S} \mathbf{P}_i) \|^2}, \quad \text{Abs.Rel.}: \frac{1}{T} \sum_{D \in T} \left|\frac{D_{i} - D_{i}^*}{D_{i}^*}\right|
%     \label{eq:metrics}
% \end{equation}
%%%%%%%%%%%%%%%%%%%%%%%%%%%%%%%%%%%%%%%%%%%%%%%%%%%%%%%%%%%%%%%%%%%%%%%%%%

In addition, we report the depth accuracy by the absolute relative error (Abs.Rel.) for further analysis.
Note that \textit{median-scaling} is applied to evaluate up-to-scale predictions unless otherwise described, but without \textit{post-process}~\cite{godard2017unsupervised}.

\subsection{Datasets}
We selected benchmarks on which DROID-SLAM exhibits significant challenges compared with established methods, specifically the ORB-SLAM series~\cite{Ye2023PVO,r3d3,yin2023metric}.

\noindent \textbf{DDAD \cite{packnet}}.
DDAD (Dense Depth for Autonomous Driving) is a dataset with various driving scenes including suburbs and busy highways. The dataset contains RGB images from six synchronized cameras, camera parameters, and dense point clouds recorded by LiDAR.
We use only front camera recorded images with downsizing into $384 \times 640$ for training ($12650$ samples) and for testing on $50$ validation sequences ($3950$ samples).
The point cloud up to $200$\si\metre\xspace is used as a valid GT for depth evaluation.

\noindent \textbf{KITTI~\cite{geiger2012we}}.
The KITTI dataset is one of the standard benchmarks for various tasks such as depth and odometry evaluation. We use the captured RGB images from the left-mounted camera in the \textit{sequence} 00--08 for self-supervised training (total $20409$ samples), and 09--10 for testing the visual odometry. All images are resized into $192 \times 640$ except for the higher resolution experiment where $288 \times 960$ is assigned (Tab.~\ref{tab:ate_kitti_hr}). Notably, the entire experiment is consistently monocular for both train and test times, unlike the method using stereo configuration for training~\cite{dfvo2020icra,doc2021icra}.
We applied the \emph{Garg} cropping~\cite{zhou2017unsupervised} and used $652$ annotated depth maps as GT~\cite{ali2019kittiano} with a range of up to $80$\si\metre to quantify the depth estimation performance.

\subsection{Implementation Details}
\label{subsec:imple}
The proposed models were implemented using PyTorch~\cite{paszke2017automatic} and trained with eight NVIDIA A10G GPUs.
ResNet18-based models~\cite{he2016resnet18} were used for both \textit{PoseNet} and \textit{DepthNet}. They were optimized with the Adam optimizer~\cite{kingma2014adam} with a learning rate of $1 \times 10^{-4}$, $100$ iteration epochs, batch size of $8$ per GPU, and color jittering that follows~\cite{tri_sesc_iros23}.
Temporally $\pm 1$ adjacent frames constructed the pair for all self-supervised experiments.
The weighting of the loss functions followed the official implementation of the SC-Depth V3~\cite{sc_depthv3_tpami_2024},
and followed Monodepth2~\cite{monodepth2} for the baseline, except for the edge-aware smoothness loss~\cite{godard2017unsupervised}: we assigned $\lambda=1 \times 10^{-4}$ for this smoothness term weighting. Each training was completed in up to approximately $11$ hours, which is shorter than up to $7$ days required for SLAM backbone fine-tuning~\cite{teed2021droid}. For the learned SLAM module, we followed the default parameters of the official
% \footnote{{\tt test\_eth3d.py} of \url{https://github.com/princeton-vl/DROID-SLAM}, \\ accessed on 04/04/2024}
DROID-SLAM~\cite{teed2021droid} demonstration code and used their off-the-shelf weights trained on TartanAir~\cite{tartanair2020iros}, except for the ablation study.
For the baseline ORB-SLAM3~\cite{orb3}, we reused the official implementation by adding the keyframe interpolation to recover the full trajectory.

In theory, the proposed approach is independent of the \zeroshot model choice, although slight performance differences might emerge. Therefore, we report two variations of the experimental results:  LeReS~\cite{Wei2021CVPRleres} guided learning (described as (L)), which is originally tested on the SC-Depth V3~\cite{sc_depthv3_tpami_2024}, and a newer up-to-scale depth predictor, Omnidata V2~\cite{kar20223d} guided one (O). 
Note that the models are all zero-shot for evaluation datasets (i.e., were not trained on them). 

\subsection{Odometry Evaluation on DDAD}\label{subsec:ddad}
\begin{figure}[t!]
    \centering
    \vspace{2.2mm}
    \begin{tabular}{@{}c@{}c@{}c@{}}
        \includegraphics[width=0.16\textwidth]{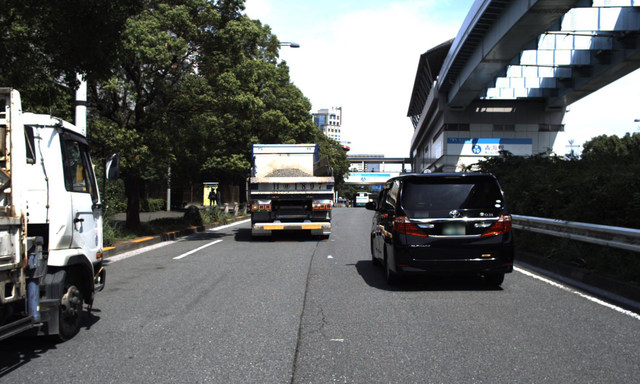}    &
        \includegraphics[width=0.16\textwidth]{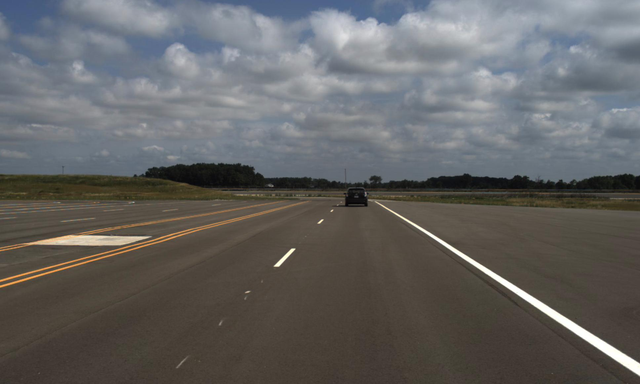}    &
        \includegraphics[width=0.16\textwidth]{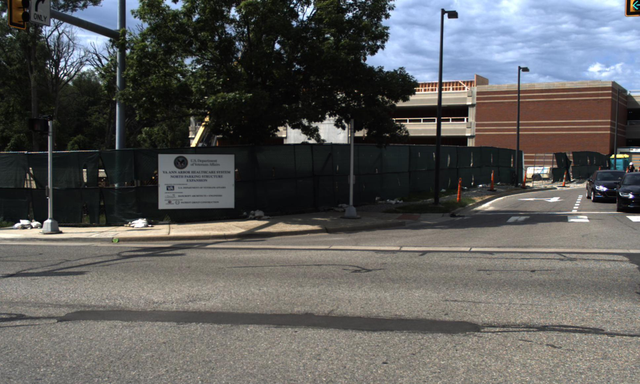}
        \\
        \includegraphics[width=0.16\textwidth]{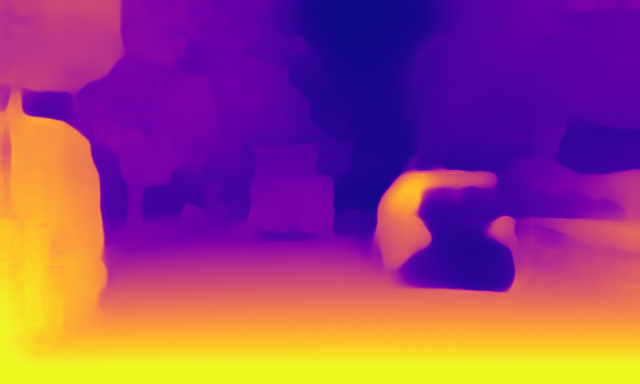}   &
        \includegraphics[width=0.16\textwidth]{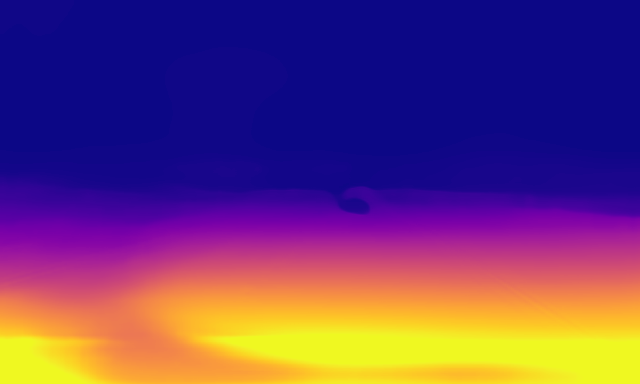}   &
        \includegraphics[width=0.16\textwidth]{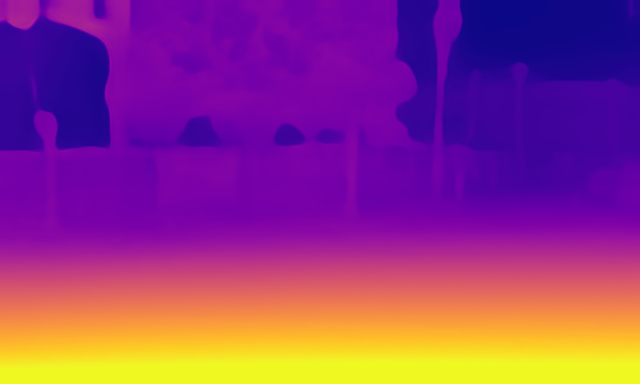}
        \\
        Abs.Rel. $0.498$                                                                                                                          & Abs.Rel. $0.120$          & Abs.Rel. $0.114$          \\ % 列名を追加
        \includegraphics[width=0.16\textwidth]{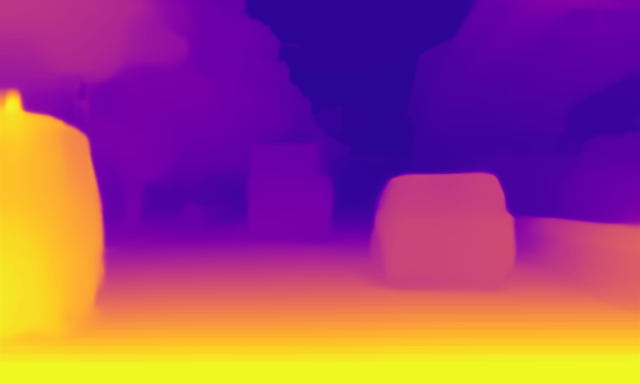} &
        \includegraphics[width=0.16\textwidth]{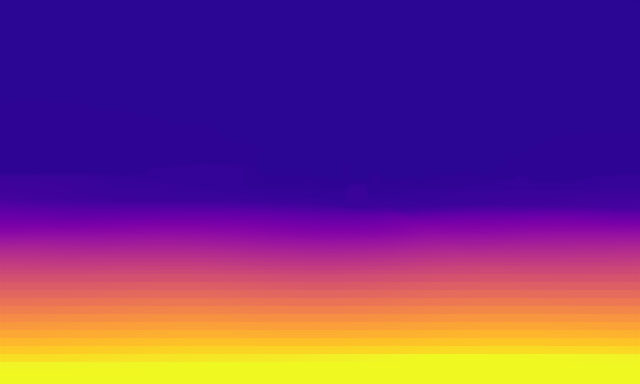} &
        \includegraphics[width=0.16\textwidth]{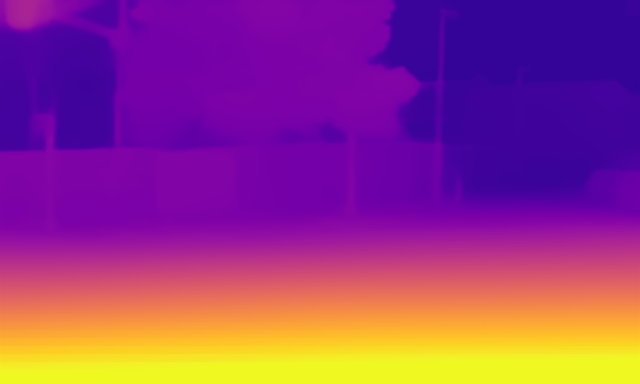}
        \\
        Abs.Rel. $0.205$                                                                                                                          & Abs.Rel. $0.101$          & Abs.Rel. $0.103$          \\ % 列名を追加
        \includegraphics[width=0.16\textwidth]{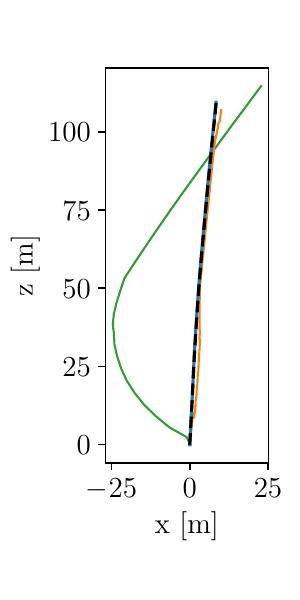}                                                 &
        \includegraphics[width=0.16\textwidth]{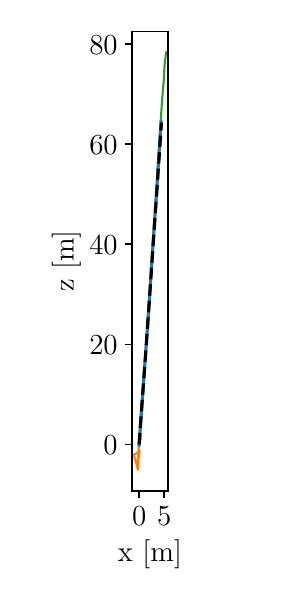}                                                 &
        \includegraphics[width=0.16\textwidth]{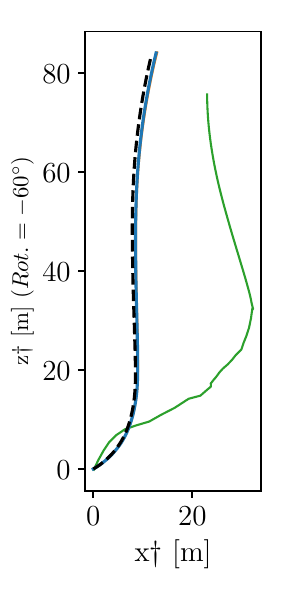}
        \\
        \multicolumn{3}{@{}c@{}}{\includegraphics[width=0.48\textwidth]{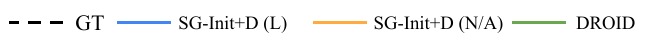}}
        \\
        (a) \textit{seq}:$000187$                                                                                                                 & (b) \textit{seq}:$000161$ & (c) \textit{seq}:$000171$ \\ % 列名を追加
    \end{tabular}
    \vspace{-1em}
    \caption{\textbf{Input/output of the visual odometry methods on DDAD.} From top to bottom, we show the input RGB image, the depth prior fed into \AcroAcro(N/A) and \AcroAcro (L), and the plotted trajectories. The reported depth prior accuracy (Abs.Rel.) is the average score through one sequence. 
    }
    \label{fig:ddad_viz_all}
    \vspace{-1em} % ADD FOR THE FIRST SUBMISSION
\end{figure}

\noindent \textbf{Performance of \Acronym Applied to DROID-SLAM.}
Table~\ref{tab:ate_ddad_mono} summarizes the result of trajectory estimation on DDAD sequences.
Our proposals that leverage both initialization (\textit{Init.}) and \zeroshot guidance (\textit{ZG}) significantly improve the estimation, even though they all depend on the same learned SLAM backbone.
For ORB-SLAM3~\cite{orb3}, we report the minimum error in five trials for each sequence, as it is stochastic. 
Figure~\ref{fig:ddad_viz_all} illustrates the benefit of depth accuracy improvement via \zeroshot guidance.
In a situation where a dynamically moving vehicle is observed (\textit{seq:}000187), where a textureless region is widely covered (\textit{seq:}000161), and where large optical flow is to be estimated because of the steep turn (\textit{seq:}000171), \zeroshot guidance demonstrates its strong contribution for the dense SLAM backbone.
% \captionsetup[table]{skip=6pt}

\begin{table}[t!]
    \vspace{2mm}
    \renewcommand{\arraystretch}{1.10}
    \centering
    \caption{
        \textbf{Trajectory estimation errors on DDAD \textit{val} split.}
    }
    {
        \small
        \setlength{\tabcolsep}{0.3em}
        \begin{tabular}{l|c|c|c|c}
            \toprule
            \textbf{Models}              &
            \textit{Init.}               &
            \textit{ZG}                  &
            \footnotesize \emph{Failure} &
            ATE $\downarrow$
            \\
            \toprule

            \textbf{\AcronymSLAM} \footnotesize{(L)}
                                         & \checkmark
                                         & \checkmark
                                         & \textbf{0}
                                         & \textbf{0.451}
            \\
            \textbf{\AcronymSLAM} \footnotesize{(O)}
                                         & \checkmark
                                         & \checkmark
                                         & \textbf{0}
                                         & \underline{0.463}
            \\

            \AcronymSLAM \footnotesize{(N/A)}
                                         & \checkmark
                                         & -
                                         & \textbf{0}
                                         & 1.152
            \\
            DROID-SLAM~\cite{teed2021droid}
                                         & -
                                         & -
                                         & \textbf{0}
                                         & 6.007
            \\
            ORB-SLAM3~\cite{orb3}
                                         & N/A
                                         & N/A

                                         & \underline{4}
                                         & 4.955
            \\
            \bottomrule
        \end{tabular}
    }
    \label{tab:ate_ddad_mono}
    \vspace{-2.em}
\end{table}

\noindent \textbf{Convergence Analysis in Challenging Scenarios.} 
Our proposal is expected to exhibit an adequate convergence in the initialization, that is, a more accurate intermediate depth prediction will be presented in its initial timestep. 
Hence, we verified its hypothesis on the success and failure scenes sampled from the four categorized groups (Fig.~\ref{fig:ddad_corr}). 
The Abs.Rel. of Figure~\ref{fig:ddad_init} shows that the estimated depth map from our proposal provides competitive accuracy on (a), along with higher accuracy even in situations (b) where a sufficient number of keyframes are challenging to obtain, (c) ego-vehicle and the scene are highly dynamic, and (d) under the moderate dynamic situation. 
Therefore, the result indicates that the proposed guidance for its initialization alleviates the convergence issue of the dense bundle adjustment.

Intriguingly, the weight prediction $\mathbf{w}_{ij}$ of our proposal exhibits a broader and more dynamic object-avoiding prediction, despite being a zero-shot prediction concerning the learned domain of the backbone.
We conjecture that the better initialization provides \emph{depth-uncertainty-awareness} to $\mathbf{w}_{ij}$, thus leading to a strong zero-shot capability in such challenging scenarios.
In fact, augmenting this weight prediction by multiplying \textit{self-discovered mask}~\cite{bian2021ijcv} --- which attempts to handle the \emph{depth-uncertainty-awareness} explicitly --- exhibits no significant improvement on ATE ($0.452$ on \AcronymSLAM (L), and $0.461$ for its ``(O)'' model).
Thus, nearly the same functionality with \textit{self-discovered mask} is expected to be obtained by our proposal.

\noindent \textbf{Comparison with Zero-shot Depth Estimators.}
As Yin \etal reported~\cite{yin2023metric}, the performance of DROID-SLAM can be improved by only providing the \zeroshot depth estimation for its initialization.
Therefore, we compare our proposal with several \zeroshot depth estimation models to answer the question, \emph{Is depth all you need?}.
The experimental result demonstrates the importance of both depth and pose initialization (Tab.~\ref{tab:ate_ddad_many_depth}), and that in which both are provided improves.
We presume that, only the accurate depth input for initialization is insufficient, and a combination of them is important because dense bundle adjustment is essentially an iterative optimization of optical flow: a combination of the depth and pose (Fig.~\ref{fig:sd3_with_dba}).

% LAbe lbox for legend
\newcommand{\redframebox}[2][\textwidth]{%
  \begin{tcolorbox}[colframe=red, width=#1, boxrule=0.3mm, colback=white, sharp corners,
      boxsep=0pt, left=3pt, right=3pt, top=1pt, bottom=1pt]
    \centering
    #2
  \end{tcolorbox}
}
\newcommand{\blackframebox}[2][\textwidth]{%
  \begin{tcolorbox}[width=#1, boxrule=0.3mm, colback=white, sharp corners,
      boxsep=0pt, left=3pt, right=3pt, top=1pt, bottom=1pt]
    \centering
    #2
  \end{tcolorbox}
}

% Size specified includegraphics with `defaultwidth`
\newcommand{\defaultwidth}{0.12}

% Function: sizeGivenGraphics
\newcommand{\sizeGivenGraphics}[2][]{%
  \begin{minipage}[b]{\defaultwidth\textwidth}
    \includegraphics[width=\textwidth]{#2}
    \if\relax\detokenize{#1}\relax
      % キャプションが指定されていない場合は何も表示しない
    \else
      \caption{#1}
    \fi
  \end{minipage}%
}

% Function: sizeGivenCaption
\NewDocumentCommand{\sizeGivenCaption}{m}{%
  \begin{minipage}[c]{\defaultwidth\textwidth}
    \centering
    \vspace{-1.em} %
    {\footnotesize #1} % テキスト
  \end{minipage}%
  % \vspace{0.1em} %
}

% Define the custom command
\newcommand{\graylabeledbox}[2][1.8cm]{%
  \colorbox{gray!20}{\makebox[#1]{#2}}%
}

\newcommand{\methodBoxSize}{2.36cm} % TODO Modify by HAND

%%%%%%%%%%%%%%%%%%%%%%%%%%%%%%% START METHOD %%%%%%%%%%%%%%%
\vspace{1.em} % Maybe 
\begin{figure}[t]
  \centering
    \vspace{2.em} % Maybe 
  \begin{tabular}{ll}
    \begin{minipage}{.07\textwidth}
      % depth-baseline
      \vspace{-.51cm}
      \rotatebox{90}{\redframebox[\methodBoxSize]{Ours}}
      \\
      \vspace{-.08cm}
      \\
      \rotatebox{90}{\blackframebox[\methodBoxSize]{Baseline}}
      \\
    \end{minipage}
    \hspace{-11mm}
    \begin{minipage}{.93\textwidth}
      \sizeGivenCaption{(a) \footnotesize{\textit{seq}:000150}}%
      \sizeGivenCaption{\blue{(b) \footnotesize{\textit{seq}:000191}}}%
      \sizeGivenCaption{\darkgreen{(c) \footnotesize{\textit{seq}:000187}}}%
      \sizeGivenCaption{\orange{(d) \footnotesize{\textit{seq}:000188}}}%
      \\
      %%%%%%%% OURS %%%%%%
      % Adding `%` just after \includeMyGraphic{} removes the space between image to image. 
      \sizeGivenGraphics{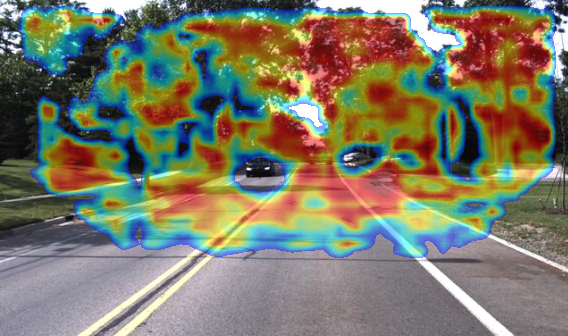}%
      \sizeGivenGraphics{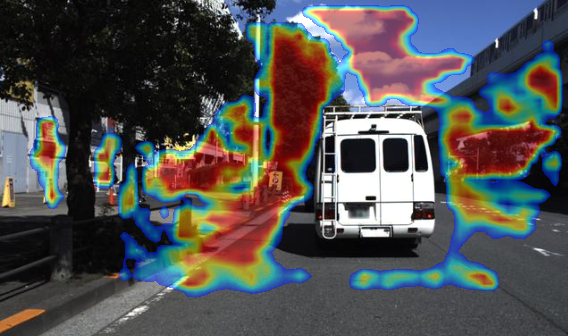}%  
      \sizeGivenGraphics{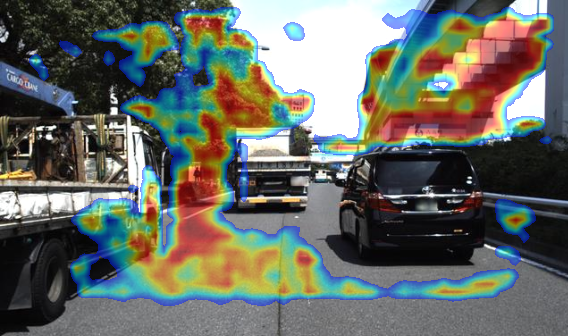}%    
      \sizeGivenGraphics{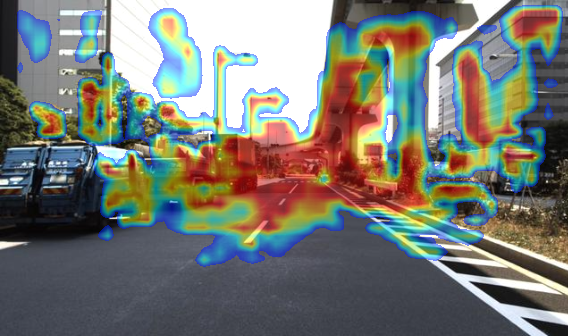}%  
      \\
      \sizeGivenGraphics{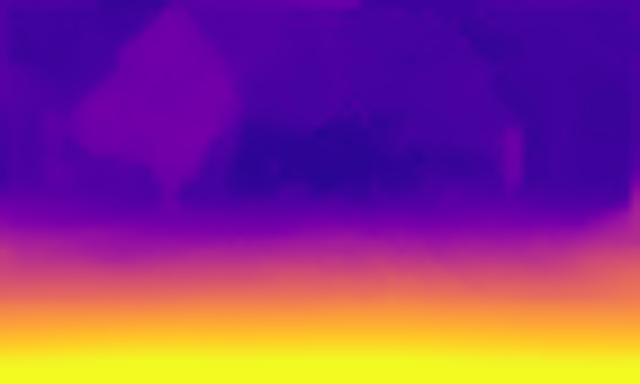}%
      \sizeGivenGraphics{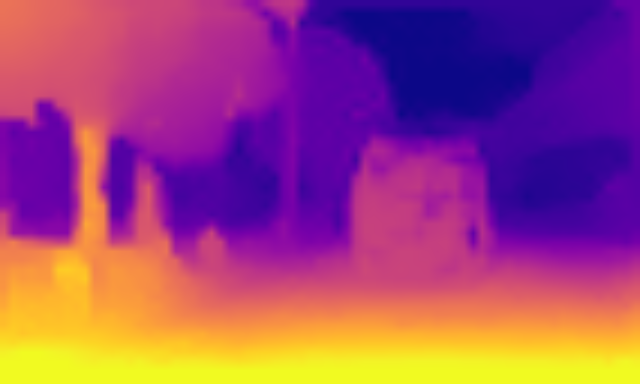}%  
      \sizeGivenGraphics{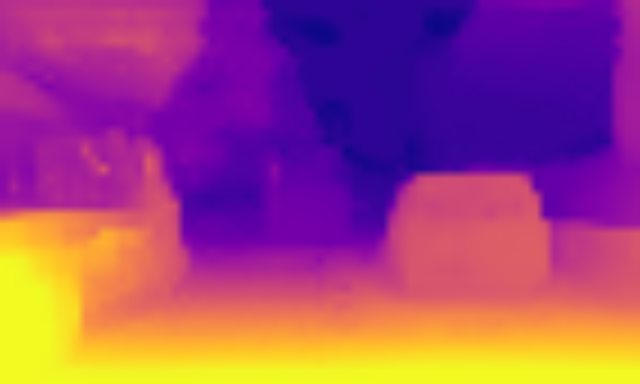}%    
      \sizeGivenGraphics{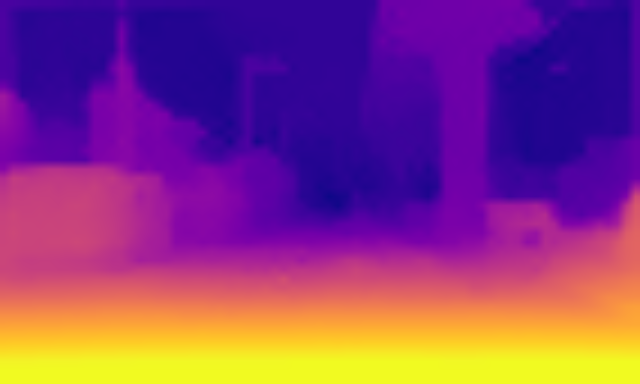}%  
      \\
      \sizeGivenCaption{$(0.124,1.388)$}%
      \sizeGivenCaption{$(0.230,0.046)$}%
      \sizeGivenCaption{$(0.240,0.499)$}%
      \sizeGivenCaption{$(0.182,0.632)$}%
      \\
      \sizeGivenGraphics{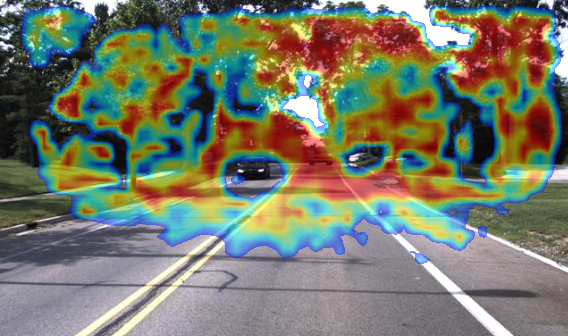}%
      \sizeGivenGraphics{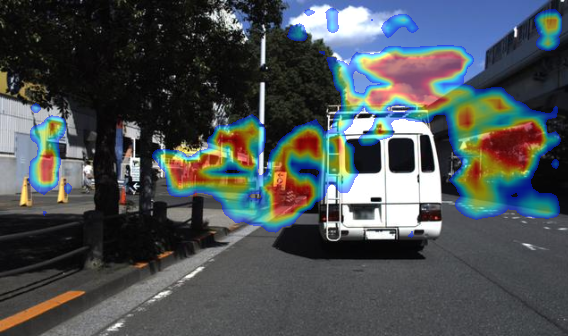}%
      \sizeGivenGraphics{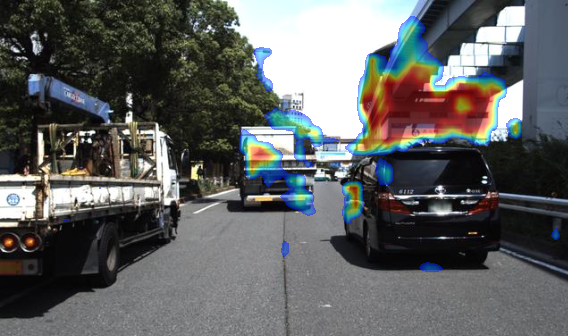}%
      \sizeGivenGraphics{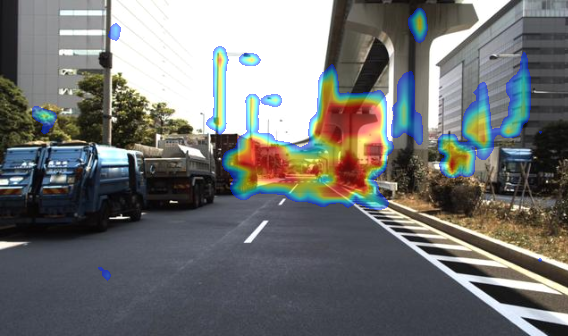}%
      \\
      % depth-baseline
      \sizeGivenGraphics{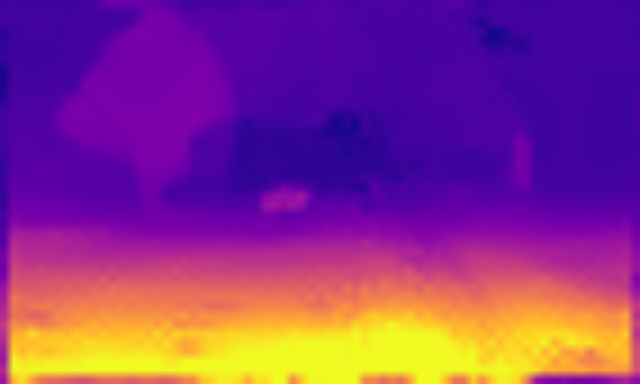}%
      \sizeGivenGraphics{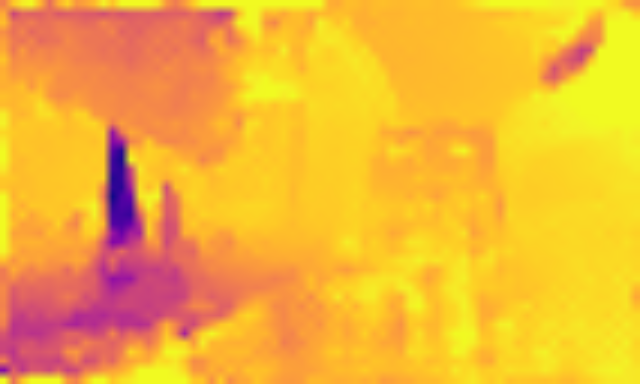}%
      \sizeGivenGraphics{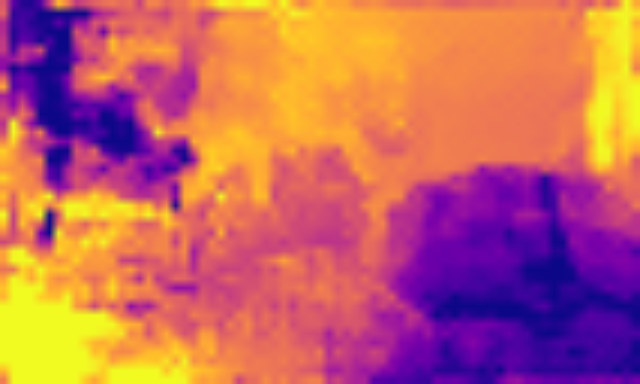}%
      \sizeGivenGraphics{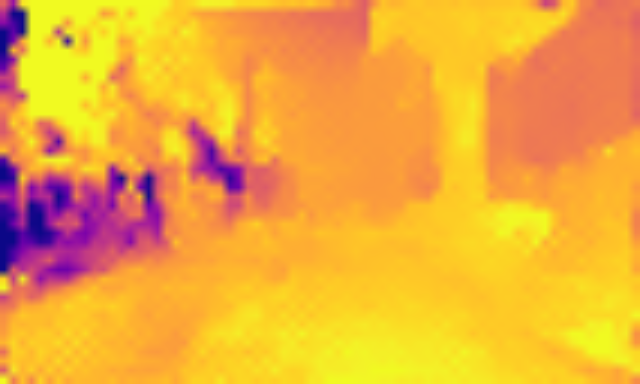}%
      \\
      \sizeGivenCaption{$(0.134,0.163)$}%
      \sizeGivenCaption{$(0.801,6.370)$}%
      \sizeGivenCaption{$(2.047,28.32)$}%
      \sizeGivenCaption{$(1.144,12.76)$}%
      \\
    \end{minipage}
  \end{tabular}
  \vspace{-2.em}
  \caption{\textbf{Analysis of the convergence with and without our proposed initialization on DDAD.} The predicted confidence weight $\mathbf{w}_{ij}$ overlayed on RGB input $I_{i}$ (top) and depth $\mathbf{d}_{i}^{-1}$ (bottom) after dense bundle adjustment are presented for each method. The tupled number represents (Abs.Rel., ATE): Abs.Rel. is calculated from this single frame. Compared to the baseline~\cite{teed2021droid}, ours generates weight that avoids moving vehicles (white ones on \blue{(b)} and a right truck and black van on \darkgreen{(c)}) and possesses a broader range in the static area. The sequence ID color corresponds to Figure~\ref{fig:ddad_corr}.}
  \label{fig:ddad_init}
  \vspace{-1.em}
\end{figure}
\begin{table}[t!]
    \vspace{2mm}
    \renewcommand{\arraystretch}{1.10}
    \centering
    \caption{
        \textbf{Trajectory estimation performance varies by depth initializer on DDAD \textit{val} split.}
        \textit{Scaling} indicates \textit{median-scaling} for MD and \textit{shift-and-scaling} for SS.
    }
    {
        \small
        \setlength{\tabcolsep}{0.3em}
        \begin{tabular}{l|c|cc}
            \toprule
            \textbf{Depth Input}         &
            \footnotesize \emph{Scaling} &
            ATE $\downarrow$             &
            Abs.Rel. $\downarrow$
            \\
            \toprule
            \multirow{2}{*}{LeReS~\cite{Wei2021CVPRleres} }
                                         & SS
                                         & 1.283
                                         & 0.274
            \\
                                         & MD
                                         & 1.174
                                         & 0.201
            \\
            \midrule
            \multirow{2}{*}{Omnidata V2~\cite{kar20223d} }
                                         & SS
                                         & 1.477
                                         & 0.300
            \\
                                         & MD
                                         & 1.347
                                         & 0.184
            \\
            \midrule
            ZeroDepth~\cite{tri-zerodepth}
                                         & -
                                         & 1.066
                                         & \underline{0.100}
            \\
            \midrule
            LiDAR Depth
                                         & -
                                         & \underline{0.908}
                                         & \textbf{0.000}
            \\
            \midrule
            \begin{tabular}{l}
                ResNet18 \\ \footnotesize{(\textbf{\AcronymSLAM})}
            \end{tabular}
                                         & MD
                                         & \textbf{0.451}
                                         & 0.143
            \\
            \bottomrule
        \end{tabular}
    }
    \vspace{-2em}
    \label{tab:ate_ddad_many_depth}
\end{table}
\noindent \textbf{Impact of the Initialization by \textit{PoseNet}.}
We ablate the \textit{PoseNet} and compare it with variants of pose estimation ways to further quantify the contribution of the \textit{PoseNet}. Table~\ref{tab:ate_ablation_ddad} summarizes the result of trajectory estimation and relationships with the depth estimation performance. Here, \AcroAcro indicates ``\AcronymSLAM'', \AcroAcroNos$\dagger$ indicates the ablation of \textit{PoseNet} from \AcroAcro, and \textit{PoseNet} describes the vanilla output of the ego-motion estimator from self-supervised learning obtained through the chaining of relative pose estimations (Eqn.~\ref{eq:camera_chain}).
The result indicates that: (1) initialization of the bundle adjustment with \textit{PoseNet} enhances trajectory estimation regardless of their depth accuracies, and (2) a model using a stronger depth estimator leads to more accurate trajectory estimation.

% \captionsetup[table]{skip=6pt}

\begin{table}[t!]
    \vspace{1em}
    \renewcommand{\arraystretch}{1.10}
    \centering
    \caption{% remove absrel && horizontal line 
    \textbf{Ablation of the \textit{PoseNet} on DDAD \textit{val} split.}
    Abs.Rel. is calculated from its \textit{DepthNet}.
    }
    {
    \small
    \setlength{\tabcolsep}{0.3em}
    \begin{tabular}{l|ccc|c}
        \toprule
        \multirow{2}{*}{\textbf{\textit{ZG} Provider}} &
        \multicolumn{3}{c|}{ATE $\downarrow$}          &
        \multirow{2}{*}{ Abs.Rel. $\downarrow$}                                                               \\
        \cmidrule{2-4}
                                                       &
        \footnotesize{\AcroAcro}
                                                       & \footnotesize{ \AcroAcroNos$\dagger$}
                                                       & \footnotesize{ \emph{PoseNet}}
                                                       &
        \\
        \toprule
        LeReS~\cite{Wei2021CVPRleres}
                                                       & \textbf{0.451}
                                                       & \underline{0.903}
                                                       & 1.674
                                                       & \textbf{0.143}
        \\
        \footnotesize{Omnidata V2}~\cite{kar20223d}
                                                       & \underline{0.463}
                                                       & \textbf{0.611}
                                                       & \underline{1.655}
                                                       & \underline{0.147}
        \\

        None
                                                       & 1.152
                                                       & 1.183
                                                       & \textbf{1.637}
                                                       & 0.195
        \\

        \bottomrule
    \end{tabular}
    }
    \vspace{-1.em}
    \label{tab:ate_ablation_ddad}
\end{table}

\noindent \textbf{Computational Efficiency.}
Finally, we compare our proposal with (1) R3D3~\cite{r3d3}, a \sota (SOTA) method that retrieves the synchronized multi-camera images, and (2) the depth-initialized model with ZeroDepth~\cite{tri-zerodepth}, a SOTA \zeroshot depth estimator (Tab.~\ref{tab:ate_ddad_vs_r3d3}).
The proposed method demonstrates competitive or superior trajectory error results, even in a monocular setup, while being approximately ten times faster and maintaining low GPU RAM usage~\footnote{The experiment is conducted on a single NVIDIA RTX A6000 with Intel Xeon Gold 6242R. Further enhancements such as model distillation and optimized implementation could improve efficiency.}.

% \captionsetup[table]{skip=6pt}

\begin{table}[t!]
 \vspace{2mm}
 \renewcommand{\arraystretch}{1.00}
 \centering
 \caption{
     \textbf{Efficiency Evaluation on DDAD \textit{val} split.} * denotes evaluated only on \textit{seq}:000150. Time is per-frame performance.
 }
 {
     \small
     \setlength{\tabcolsep}{0.3em}
     \begin{tabular}{l|ccc}
         \toprule
         \textbf{Models} &
         ATE$\downarrow$ \lbrack m\rbrack &
         Time*$\downarrow$ \lbrack s\rbrack &
         V-RAM*$\downarrow$ \lbrack GB\rbrack
         \\
         \midrule
         \textbf{\AcronymSLAM}
         & \underline{0.451}
         & \textbf{0.32}
                                               & \textbf{10.1}
         \\
         % % \toprule
         {\footnotesize ZeroDepth~\cite{tri-zerodepth} + DROID}
                                               & 1.066
                                               & \underline{1.2}
                                               & \underline{12.6}
         \\
         % \midrule
         % % \toprule
         R3D3~\cite{r3d3}
                                               & \textbf{0.433}
                                               & 3.2
                                               & 18.8
         \\
         \bottomrule
     \end{tabular}
 }
 \label{tab:ate_ddad_vs_r3d3}
 \vspace{-2em}
\end{table}

\subsection{Odometry Evaluation on KITTI Benchmark}\label{subsec:kitti}
\noindent \textbf{Evaluation of Standard Configurations.}
Table~\ref{tab:ate_kitti} summarizes the visual odometry result on KITTI.
Our proposal of \zeroshot depth-guided methods achieved competitive results with strong baselines~\cite{pan2020pseudoegbd,dfvo2020icra}.
Therefore, we can emphasize that in the conventional benchmark, unlike DDAD~\cite{packnet} in which it was relatively filled with a static observation~\cite{sc_depthv3_tpami_2024} and no failure on the keyframe detection ways was observed~\cite{murORB2, orb3}, our proposal still achieves competitive results compared with SOTA methods.
In addition, even in a monocular setup for the self-supervised training stage, the proposed approach demonstrates better trajectory estimation by leveraging the more accurate depth prediction and the consequent dense bundle adjustment, particularly in the \textit{Seq:}09 where traditional strategies are considered to suffer scale drift problems~\cite{doc2021icra}.
Note that, the ablation of our proposal, ``(N/A) version'' is inferior to the method in which self-supervised prior is \emph{sparsely} fed into the SLAM backbone~\cite{pan2020pseudoegbd}. We presume that the local degradation of self-supervised depth estimation harmed the global performance of trajectory estimation. 

However, contrary to the DDAD experiment, the accuracy of the trajectory estimation does not completely follow the order of depth estimation accuracy (Tab.~\ref{tab:kitti_depth}).
We hypothesized that it emerged from the miss-prediction on the depth map where GT evaluation was unavailable. 
Because the evaluative zone is only limited to the range of LiDAR recorded and its surrounding areas, completely quantifying the error of all pixels on the depth maps is impossible. Qualitative results suggest the effect of non-evaluative areas for depth (Fig.~\ref{fig:kitti_quality}): contrary to the ``no zero-shot depth leveraged'' model, the model with \zeroshot guidance exhibits less significant artifacts on the sky and trees. As all pixels are input into a dense bundle adjustment module, that area can be a noise for optimization.

% Size specified includegraphics with `defaultwidth`
\newcommand{\defaultwidthKitti}{0.16}

% Function: sizeGivenGraphics
\newcommand{\sizeGivenGraphicsKitti}[2][]{%
    \begin{minipage}[b]{\defaultwidthKitti\textwidth}
        \includegraphics[width=\textwidth]{#2}
        \if\relax\detokenize{#1}\relax
            % キャプションが指定されていない場合は何も表示しない
        \else
            \caption{#1}
        \fi
    \end{minipage}%
}

% Function: sizeGivenCaption
\NewDocumentCommand{\sizeGivenCaptionKitti}{m}{%
  \begin{minipage}[c]{\defaultwidthKitti\textwidth}
    \centering
    \vspace{-1.em} %
    {\footnotesize #1} % テキスト
  \end{minipage}%
  % \vspace{0.1em} %
}

\begin{figure}[t!]
    \centering
    \vspace{2.2mm}
    \begin{tabular}{@{}c@{}c@{}c@{}c@{}}
        \sizeGivenGraphicsKitti{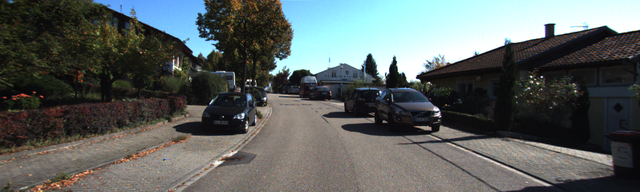}%
        % \sizeGivenGraphics{images/kitti_depths/544/media_images_val-KITTI_odometry-0910-cam0-544-pred-depth_-0_0-_0_0_0544LeReS.png}%
        \sizeGivenGraphicsKitti{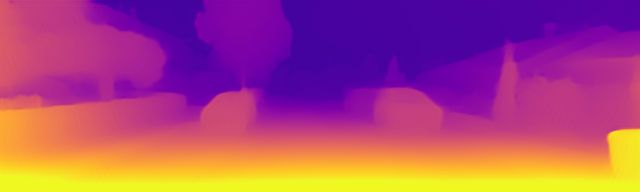}%
        \sizeGivenGraphicsKitti{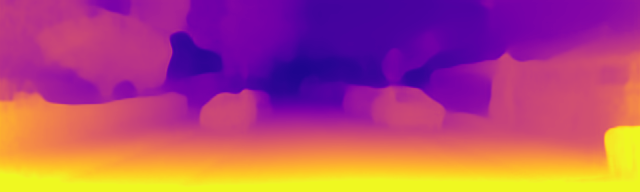}
        \\
        % \vspace{-3mm}
        %%%%%%%%%%%%%%%%%%%%%%%%%%%%%%%%%%%%%%%%%%%%%%%%%
        \sizeGivenGraphicsKitti{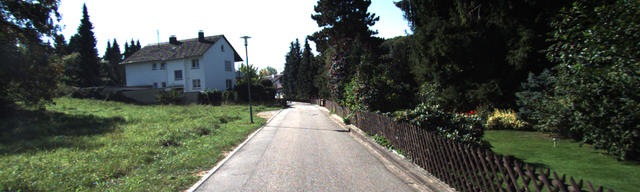}%
        % \sizeGivenGraphics{images/kitti_depths/1664/media_images_val-KITTI_odometry-0910-cam0-1664-pred-depth_-0_0-_0_0_LeReS.png}%
        \sizeGivenGraphicsKitti{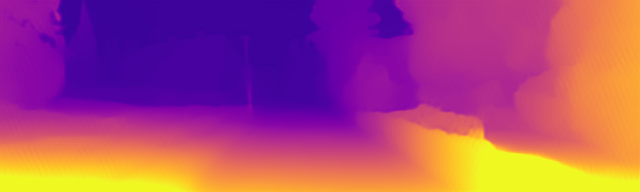}%
        \sizeGivenGraphicsKitti{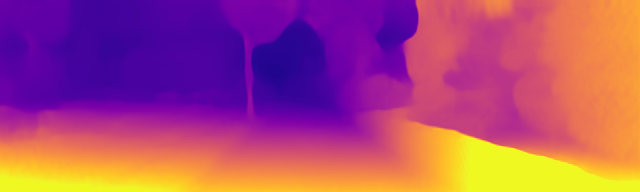}
        \\
        \sizeGivenCaptionKitti{Input}%
        % \sizeGivenCaption{LeReS}%
        \sizeGivenCaptionKitti{Omnidata V2}%
        \sizeGivenCaptionKitti{N/A}%
        \\
    \end{tabular}
    \caption{\textbf{Learned depth input for \Acronym.} The top shows the result of \textit{seq:}$09$ and \textit{seq:}$10$ below.}
    \label{fig:kitti_quality}
    \vspace{-1em}
\end{figure}

\begin{table}[t!]
    % \vspace{2mm}
    \renewcommand{\arraystretch}{1.10}
    \centering
    \caption{
        \textbf{Trajectory estimation result on KITTI benchmark.}
    }
    {
        \small
        \setlength{\tabcolsep}{0.3em}
        \begin{tabular}{l|ccc}
            \toprule
            \multirow{2}{*}{\textbf{Models}} &
            \multicolumn{3}{c}{ATE$\dagger$ $\downarrow$}                      \\
            \cmidrule{2-4}
                                             & \xspace \textit{Seq}:09 \xspace
                                             & \xspace \textit{Seq}:10 \xspace
                                             & \xspace \textit{Ave.} \xspace
            \\
            \toprule
            \textbf{\Acronym + DROID} \footnotesize{(L)}
                                             & \textbf{8.37}
                                             & 9.76
                                             & \underline{9.07}
            \\
            \textbf{\Acronym + DROID} \footnotesize{(O)}
                                             & \underline{8.64}
                                             & 10.14
                                             & 9.39
            \\
            % \midrule

            \Acronym + DROID \footnotesize{(N/A)}
                                             & 19.08
                                             & 10.77
                                             & 14.92                           % (19.08+10.77)/2.
            \\
            DROID-SLAM~\cite{teed2021droid}
                                             & 77.73
                                             & 15.87
                                             & 46.80                            % (77.73+15.87)/2.
            \\
            ORB-SLAM3~\cite{orb3}
                                             & 64.74
                                             & 80.17
                                             & 72.45
            \\
            \midrule
            \midrule
            DF-VO \footnotesize{(Mono-SC Train)}~\cite{dfvo2020icra}
                                             & 11.02
                                             & \textbf{3.37}
                                             & \textbf{7.20}                   % (11.02+3.37)/2=7.195
            \\
            pRGBD-Refined~\cite{pan2020pseudoegbd} % (11.97+6.35)/2=
                                             & 11.97
                                             & \underline{6.35}
                                             & 9.16
            \\
            PVO~\cite{Ye2023PVO}
                                             & 14.65
                                             & 8.66
                                             & 11.66                           % (14.65 + 8.66)/2 = 11.655
            \\
            \bottomrule
        \end{tabular}
    }
    \label{tab:ate_kitti}
    \vspace{-1em}
\end{table}

\begin{table}[t!]
    \vspace{2mm}
    \renewcommand{\arraystretch}{1.10}
    \centering
    \caption{
        \textbf{Depth estimation results on KITTI annotated split~\cite{ali2019kittiano}.}
        }
    {
        \small
        \setlength{\tabcolsep}{0.3em}
        \begin{tabular}{l|cccc}
            \toprule
            \textbf{\textit{ZG} Provider} &
            Abs.Rel.$\downarrow$           &
            Sq.Rel.$\downarrow$            &
            RMSE$\downarrow$              &
            $\delta_{1.25}$ $\uparrow$
            \\
            \toprule
            LeReS~\cite{Wei2021CVPRleres}
                                          & \textbf{0.090}
                                          & \textbf{0.482}
                                          & \textbf{3.974}
                                          & \textbf{0.907}
            \\
            Omnidata V2~\cite{kar20223d}
                                          & 0.103
                                          & 0.612
                                          & 4.647
                                          & 0.876
            \\
            None
                                          & \underline{0.094}
                                          & \underline{0.551}
                                          & \underline{4.063}
                                          & \underline{0.900}
            \\
            \bottomrule
        \end{tabular}
    }
    \label{tab:kitti_depth}
    \vspace{-1em}
\end{table}

\noindent \textbf{High-Resolution Experiment.}
Next, we summarize the result of the higher-resolution version in Table~\ref{tab:ate_kitti_hr} to eliminate the potential issue that a feature-matching-based strategy suffers from the reduced number of matching candidates by the downsizing of the input image.
To verify this, we feed the image of the original resolution ($376 \times 1241$) into ORB-SLAM3~\cite{orb3}, and feed $288 \times 960$ to train the \textit{DepthNet} and \textit{PoseNet} for our proposal. Note that its size maintains the same learning configuration by GPUs with $24$ GB memory, as described in Section~\ref{subsec:imple}.
The result demonstrates that ORB-SLAM3~\cite{orb3} obtains a more accurate result than the scores listed in Table~\ref{tab:ate_kitti} by a higher resolution.
Despite that improvement, the proposed approach still achieves better results in this configuration.

\subsection{Performance Difference between SLAM Backbones}
Finally, we investigated whether the proposed initialization enhances the learned SLAM independent of what their backbone learned. For the comparison, we selected the VKITTI2~\cite{cabon2020vkitti2} learned backbone as it is usually applied for previous works~\cite{Ye2023PVO,r3d3}. As summarized in Table~\ref{tab:vkitti_taratan}, our proposal demonstrates its contribution regardless of fine-tuning dataset.
Additionally, the result can be understood as evidence of less contribution to the ``seemingly in-domain'' fine-tuning of the backbone than the proposed initialization. We conjecture that rather than fine-tuning on synthetic data that mimics the domain of the real environment to adapt, leveraging the large-scale pre-trained backbone efficiently is preferable in this learning-based scheme.

%%% More concretely mentioned about the limitation on indoor
\section{Limitations}
Because \Acronym relies on dense bundle adjustment, it still requires a larger memory footprint than traditional strategies.
In addition, our proposal assumes that self-supervised depth and ego-motion models are correctly learned. 
Therefore, applying the proposed method to scenarios with varying depth ranges and significant rotational motions (i.e., indoors) may degrade odometry performance~\cite{runze2023monoindoor}.
A study to realize the best accuracy in general settings (e.g., indoors) while maintaining the capability against scenes that we verified in this work (driving) is an interesting research question.

\begin{table}[t!]
    \vspace{2mm}
    \renewcommand{\arraystretch}{1.10}
    \centering
    \caption{
        \textbf{Trajectory estimation result on KITTI benchmark with High Resolution.} 
        ``MR" is the result from $192 \times 640$ input.
    }
    {
        \small
        \setlength{\tabcolsep}{0.3em}
        \begin{tabular}{l|ccc}
            \toprule
            \multirow{2}{*}{\textbf{Models}} &
            \multicolumn{3}{c}{ATE$\dagger$ $\downarrow$}                      \\
            \cmidrule{2-4}
                                             & \xspace \textit{Seq}:09 \xspace
                                             & \xspace \textit{Seq}:10 \xspace
                                             & \xspace \textit{Ave.} \xspace
            \\
            \toprule
            \textbf{\AcronymSLAM} \footnotesize{(L)}
                                             & \underline{8.46}
                                             & \textbf{7.05}
                                             & \textbf{7.76}                   % (8.458+7.052)/2
            \\
            \textbf{\AcronymSLAM} \footnotesize{(O)}
                                             & 8.72
                                             & 8.70
                                             & 8.71                            % (8.458+7.052)/2
            \\
            ORB-SLAM3 ~\cite{orb3}
                                             & 8.61
                                             & \underline{7.73}
                                             & \underline{8.19}
            \\
            \midrule
            \midrule
            \textbf{\AcronymSLAM} \footnotesize{(L, MR)}
                                             & \textbf{8.37}
                                             & 9.76
                                             & 9.07
            \\
            \bottomrule
        \end{tabular}
    }
    % \vspace{-3mm}
    \label{tab:ate_kitti_hr}
\end{table}
\begin{table}[t!]
    % \vspace{2mm}
    \renewcommand{\arraystretch}{1.10}
    \centering
    \caption{
        \textbf{Improvement by our proposal on various SLAM backbones.} See Section~\ref{subsec:ddad} and ~\ref{subsec:kitti} for the metrics.
    }
    {
        \small
        \setlength{\tabcolsep}{0.3em}
        \begin{tabular}{l|l|c|c}
            \toprule
            \multirow{2}{*}{\textbf{Fine-tuning}} &
            \multirow{2}{*}{Model}           &
            \multicolumn{2}{c}{Benchmarks}                                        \\
            \cmidrule{3-4}
                                              &
                                              & \xspace DDAD~\cite{packnet}
                                              & \xspace KITTI~\cite{geiger2012we}
            \\
            \toprule
            \multirow{2}{*}{ N/A }
                                              & \footnotesize{DROID-SLAM}
                                              & 6.007
                                              & 46.8
            \\
                                              & \textbf{\AcroAcro}
                                              & \textbf{0.451}
                                              & \textbf{9.07}
            \\
            \midrule
            \multirow{2}{*}{VKITTI2~\cite{cabon2020vkitti2} }
                                              & \footnotesize{DROID-SLAM}
                                              & 14.21
                                              & 31.2
            \\
                                              & \textbf{\AcroAcro}
                                              & \underline{1.031}
                                              & \underline{9.77}
            \\
            \bottomrule
        \end{tabular}
    }
    \label{tab:vkitti_taratan}
    \vspace{-1em}
\end{table}

%%%%%%%%%%%%%%%%%%%%%%%%%%%%%%%%%%%%%%%%%%%%%%%%%%%%%%%
\section{Conclusion}
In this study, we investigated the strengths and weaknesses of learning-based SLAM with dense bundle adjustment, and evaluated its potential for robustness and generalizability in driving scenarios.
We found that proper handling of the nonlinear optimization is crucial for accurate trajectory estimation in this setting by analyzing the failure cases.
We propose a novel initialization strategy, \Acronym, that leverages the self-supervised depth and ego-motion learning principle combined with a large-scale pre-trained depth estimator to initialize the nonlinear optimization.
A comprehensive analysis of our proposal in a real-world outdoor driving environment revealed the benefit of this approach: our initialization enables trajectory estimation that is competitive with SOTA methods without any further training of the SLAM backbone.

%%%%%%%%%%%%%%%%%%%%%%%%%%%%%%%%%%%%%%%%%%%%%%%%%%%%%%%
\section{Acknowledgments}
We are grateful to Kota Shinjo, Dr. Shigemichi Matsuzaki, Dr. Shintaro Yoshizawa, Yuto Mori, and Dr. Rares Ambrus for their outstanding effort and enthusiasm in this study.
In the implementation phase, this study was supported by OpenAI's ChatGPT-3.5.
We would like to thank Editage (www.editage.jp) for English language editing.

% %%%%%%%%%%%%%%%%%%%%%%%%%%%%%%%%%%%%%%%%%%%%%%%%%%%%%%%
\ifarxiv
  
 % Just for arXiv submission
\else
  \bibliographystyle{IEEEtran}
  \bibliography{references}
\fi

\end{document}